\documentclass[pmlr,twocolumn,10pt]{jmlr}

\usepackage{booktabs}
\usepackage{siunitx}

\usepackage[switch]{lineno}

\usepackage{pifont}
\usepackage{multirow}
\usepackage{placeins}

\theorembodyfont{\upshape}
\theoremheaderfont{\scshape}
\theorempostheader{:}
\theoremsep{\newline}

\jmlrvolume{333}
\jmlryear{2026}
\jmlrpages{404--426}
\jmlrsubmitted{LEAVE UNSET}
\jmlrpublished{LEAVE UNSET}
\jmlrworkshop{Conference on Health, Inference, and Learning (CHIL) 2026}

\title[Benchmarking Stepwise Diagnostic Reasoning]{ASCENT: A Benchmark for Evaluating and Advancing \titlebreak Stepwise Diagnostic Reasoning in Large Language Models on Common Clinical Scenarios}

\author{%
\Name{Yera Choi} \Email{yera.choi@navercorp.com}\\
\addr NAVER Applied AI Research, Republic of Korea
\AND
\Name{Yeong Hwa Kim}\thanks{Work done while at NAVER Applied AI Research (formerly NAVER Digital Healthcare Lab).} \Email{yhkimmay@ewhain.net}\\
\addr Ewha Womans University Mokdong Hospital, Republic of Korea
\AND
\Name{JaeDeok Lee}\footnotemark[1] \Email{jaedeok.d.lee@gmail.com}\\
\addr NAVER Applied AI Research, Republic of Korea\\
Healthcare AI Research Institute (HARI), Seoul National University Hospital, Republic of Korea
\AND
\Name{Taekang Kim}\footnotemark[1] \Email{taekang@yonsei.ac.kr}\\
\addr Yonsei University Wonju College of Medicine, Republic of Korea
\AND
\Name{Sangdoo Yun} \Email{sangdoo.yun@navercorp.com}\\
\addr NAVER AI Lab, Republic of Korea
\AND
\Name{Seong-Eun Moon}\footnotemark[1] \Email{seongeun.moon89@gmail.com}\\
\addr NAVER Applied AI Research, Republic of Korea
}

\begin{document}
\firstpageno{404}

\maketitle

\begin{abstract}
Large language models (LLMs) excel at medical question answering yet are rarely evaluated on the \emph{stepwise} diagnostic reasoning that defines real clinical workflows, where impressions are revised as information accumulates. We build \textbf{Annotated Stepwise Clinical rEasoning for NaturalisTic Diagnosis (ASCENT)}, a clinician-annotated benchmark and training resource of 3{,}078 stepwise problems derived from MedQA-USMLE that decomposes each vignette into EMR-aligned steps (Findings, Impression, supporting Rationale), enabling evaluation of intermediate reasoning under incomplete information.

Experiments and training with ASCENT revealed insights into how current LLMs handle stepwise diagnostic reasoning. Even strong reasoning models that perform well on MedQA-USMLE leave substantial headroom on ASCENT, and general-purpose frontier models trail further—exposing a persistent gap between fully informed and stepwise diagnosis. Fine-tuning Qwen2.5-7B and 32B on ASCENT yields measurable F1 gains over both pre-trained and HuatuoGPT-o1 CoT-trained baselines, with gains driven primarily by precision. Complementary robustness analyses (counterfactual perturbation, format-vs-content control, judge agreement, and rollout) further show that ASCENT-fine-tuned models rely on the diagnostic content of prior impressions rather than imitating their output format, while error propagation under rollout remains a key challenge for clinical deployment.
\end{abstract}

\paragraph*{Data and Code Availability}
This paper uses the MedQA-USMLE benchmark dataset, which is available on the HuggingFace Datasets repository \citep{Jin2020MedQAUSMLE}.
The proposed dataset and code accompanying the paper will be made publicly available\footnote{The data and code will be released at \url{https://github.com/naver-ai/ascent}.}.

\paragraph*{Institutional Review Board (IRB)}
The current research does not require IRB approval, as it utilized only publicly available medical license examination materials.

\section{Introduction}
\label{sec:intro}

Recently, large language models (LLMs) have been progressively utilized in the medical domain and have demonstrated remarkable performance on tasks including medical question answering (QA) \citep{Singhal2023TowardsEM} and differential diagnosis \citep{Savage2023DiagnosticRP}. With the advent of reasoning models that leverage test-time scaled Chain-of-Thought (CoT) prompting \citep{Wei2022CoT} and effective learning strategies incorporating reinforcement learning (RL), LLMs are even considered capable of generating complex clinical reasoning and solving complex diagnostic problems \citep{Chen2024HuatuoGPTo1TM}.

Nevertheless, the application of model-generated clinical reasoning outcomes to real-world clinical settings has been limited, which is in part due to the difficulty of assessing their clinical validity and effectiveness \citep{Meng2024TheAO}. Unlike mathematical or coding problems, which generally allow logical step-by-step formulation and automatic verification, medical problems are difficult to convert into clear step-by-step structures and require expert involvement for rigorous evaluation of solutions \citep{Wu2025MedReasonEF}.

Accordingly, although still in the early stages, existing works have attempted to ground the generation of clinical reasoning through LLMs, medical knowledge resources, or human experts, and increasingly emphasize the importance of stepwise verification to create valid clinical reasoning paths. HuatuoGPT-o1 and ReasonMed rely on LLM-based medical verifiers \citep{Chen2024HuatuoGPTo1TM, Sun2025ReasonMedA3}, while MedReason implements a medical knowledge graph to create reasoning chains \citep{Wu2025MedReasonEF}. MedCaseReasoning focuses on clinician–model alignment in case studies \citep{Wu2025MedCaseReasoningEA}.

A more recent study, SD Bench \citep{Nori2025SequentialDW}, highlights a common shortcoming of prior approaches--their reliance on complete patient vignettes that include all available information for diagnosis. Such approaches tend to overlook the importance of ``sequential diagnosis,'' a crucial component of real-world clinical decision-making \citep{Daniel2019ClinicalRA}. While this study draws important attention to the need for stepwise reasoning, it, along with other recent works, primarily focuses on complex and challenging case studies. Although interpreting such difficult cases is valuable, such a focus may lead to overlooking the need to assess models’ capabilities in commonly encountered scenarios. Moreover, SD Bench does not explicitly evaluate intermediate reasoning outputs under uncertainty, leaving LLMs' stepwise reasoning capacity underexplored.

To support the development and evaluation of stepwise clinical reasoning, we propose \textbf{Annotated Stepwise Clinical rEasoning for NaturalisTic Diagnosis (ASCENT)}, a clinician-annotated dataset derived from the widely used MedQA-USMLE benchmark \citep{Jin2020MedQAUSMLE}. ASCENT targets a gap that prior datasets leave open: although diagnostic reasoning is crucial for accurate clinical decision-making, its core cognitive processes are typically acquired implicitly rather than explicitly taught \citep{Graber2005DiagnosticEI}, so expert reasoning in common scenarios is rarely articulated in a stepwise manner—precisely the regime LLMs need to learn.

ASCENT structures each case incrementally along sections commonly found in Electronic Medical Records (EMRs)—such as Present Illness and Physical Examination—approximating the natural flow of information at the bedside. At every step, clinicians annotate the Findings, the Impression (suspected disorders given the information so far), and the Reason (the rationale supporting the Impression). This dual annotation of both structured diagnostic targets and rationales enables fine-grained, intermediate evaluation that goes beyond final-answer correctness, and—because ASCENT inherits from MedQA-USMLE—covers a broad range of commonly encountered, representative clinical scenarios that are often underexplored despite forming the basic foundation of medical reasoning.

We further demonstrate that training on ASCENT yields substantial empirical gains. Models ranging from 7B to 32B in size exhibit an F1 score improvement of 0.05--0.11 over their pre-trained baselines. Moreover, compared to models trained on complex CoT annotations from HuatuoGPT-o1, models trained on ASCENT outperform by 0.04--0.11 F1, suggesting that ASCENT effectively complements existing datasets by recovering basic but essential reasoning steps that are often overlooked.

The key contributions of the current study are as follows:
\begin{itemize}
    \item We build and release \textbf{ASCENT}, a benchmark dataset of structured patient vignettes segmented into \textbf{intermediate diagnostic steps} with \textbf{stepwise evaluation targets}. Each step includes clinician-verified findings, suspected impressions, and supporting rationales, offering a transparent view of the stepwise clinical reasoning process.

    \item ASCENT reflects the structure of real-world EMR sections (Present Illness, Physical Examination, and so on), approximating the natural flow of information in clinical settings.

    \item Unlike previous datasets that focus primarily on complex or rare cases, ASCENT emphasizes \textbf{common and exemplar clinical scenarios}, which are often governed by implicit reasoning and underrepresented in prior work.

    \item By reconstructing these implicit decisions into explicit, interpretable reasoning chains, ASCENT offers a framework to enhance and evaluate stepwise diagnostic reasoning in LLMs.
\end{itemize}

\section{Related Work}
\paragraph{LLM-based clinical reasoning and differential diagnosis.}
LLMs have shown increasing potential for clinical reasoning, with early efforts exploring how specialized optimization can enhance diagnostic performance. \textbf{Articulate Medical Intelligence Explorer (AMIE)} \citep{McDuff2025AMIE} represents one of the first attempts to tailor LLMs for diagnostic reasoning, demonstrating strong performance on 302 challenging case-study problems. \citet{Brodeur2024SuperhumanPO} report a complementary finding that a frontier general-purpose LLM matches or exceeds physician performance on standardized diagnostic-reasoning tasks, further motivating fine-grained evaluation of how this capability transfers to stepwise clinical workflows.

Subsequent research has advanced long-form medical reasoning in LLMs by improving reliability and stepwise explainability. \textbf{HuatuoGPT-o1} \citep{Chen2024HuatuoGPTo1TM} built 40K verifiable exam-style problems and applied supervised fine-tuning (SFT) with verifier-based reinforcement learning with verifiable rewards (RLVR), while \textbf{m1} \citep{Huang2025m1UT} showed that test-time scaling alone can boost sub‑10B models to rival larger ones, though excessive “overthinking” can hurt accuracy. Complementary efforts such as \textbf{MedReason} \citep{Wu2025MedReasonEF} and \textbf{ReasonMed} \citep{Sun2025ReasonMedA3} focus on intermediate reasoning chains—via knowledge‑graph paths or multi‑agent verification—yielding improved faithfulness and enabling sub‑10B models to approach or surpass much larger models. Together, these studies highlight the central role of curated step‑by‑step reasoning data in developing trustworthy medical LLMs.

Complementing training‑focused efforts, recent benchmarks emphasize sequential diagnosis and clinician‑aligned reasoning. \textbf{SD Bench} \citep{Nori2025SequentialDW} transforms 304 cases from the New England Journal of Medicine Clinicopathological Conferences (NEJM-CPC) into interactive stepwise encounters; \textbf{MedCaseReasoning} \citep{Wu2025MedCaseReasoningEA} offers 14K QA cases with clinician‑authored reasoning to assess diagnostic accuracy and reasoning fidelity; \textbf{MedThink-Bench} \citep{Zhou2025MedThinkBench} provides 500 expert‑annotated cases with an LLM‑as‑a‑Judge framework for fine‑grained evaluation; and \textbf{HealthBench} \citep{Arora2025HealthBenchEL} introduces 5,000 multi‑turn conversations evaluated against physician‑authored rubrics for open‑ended healthcare interactions, including settings where the model must elicit missing patient information through follow‑up questions, although its scope spans a broader range of healthcare interactions. While these efforts substantially broaden evaluation coverage, no existing benchmark systematically evaluates intermediate reasoning steps under conditions of clinical uncertainty in common diagnostic scenarios. Table \ref{tab:data_comparison} compares datasets focused on medical reasoning within diagnostic scenarios.

\begin{table}[t]
  \centering
  \setlength{\tabcolsep}{1mm}
  \small
   \begin{tabular}{lcccc}
    \toprule
    \textbf{Dataset} & \multicolumn{2}{c}{\textbf{Expert}} & \multicolumn{1}{c}{\textbf{GT}} & \multicolumn{1}{l}{\textbf{Intermediate}} \\
     & \multicolumn{1}{c}{\textbf{Review}} &  & \multicolumn{1}{c}{\textbf{Rationale}} & \multicolumn{1}{c}{\textbf{Target}} \\
    \midrule

    HuatuoGPT-o1 & \textcolor{red}{\ding{55}} &  & \textcolor{green}{\ding{51}} & \textcolor{red}{\ding{55}} \\
    m1         & \textcolor{red}{\ding{55}} &  & \textcolor{green}{\ding{51}} & \textcolor{red}{\ding{55}} \\
    MedReason & \textcolor{orange}{$\triangle$} &  & \textcolor{green}{\ding{51}} & \textcolor{red}{\ding{55}} \\
    ReasonMed & \textcolor{red}{\ding{55}} &  & \textcolor{green}{\ding{51}} & \textcolor{red}{\ding{55}} \\
    MedCaseReasoning & \textcolor{green}{\ding{51}} &  & \textcolor{green}{\ding{51}} & \textcolor{red}{\ding{55}} \\
    SD Bench & \textcolor{green}{\ding{51}} &  & \textcolor{red}{\ding{55}} & \textcolor{red}{\ding{55}} \\
    MedThink-Bench & \textcolor{green}{\ding{51}} &  & \textcolor{green}{\ding{51}} & \textcolor{red}{\ding{55}} \\
    \texttt{ASCENT} (ours) & \textcolor{green}{\ding{51}} &  & \textcolor{green}{\ding{51}} & \textcolor{green}{\ding{51}} \\
    \bottomrule
  \end{tabular}
  \caption{Comparison between datasets. \textbf{Expert Review}: Whether the dataset’s annotations were reviewed by medical experts. \textbf{GT Rationale}: Whether the dataset provides ground‑truth explanations.
  \textbf{Intermediate Target}: Whether the dataset includes intermediate reasoning targets for sequential diagnostic evaluation.
  In the case of MedReason, \textcolor{orange}{$\triangle$} under Expert Review indicates partial evaluation, where over 25 sampled questions across seven specialties were evaluated by directly comparing anonymized HuatuoGPT‑o1 and MedReason Chain-of-Thought data.}
  \label{tab:data_comparison}
\end{table}

\begin{figure*}[t]
\centering
\includegraphics[width=1.05\textwidth]{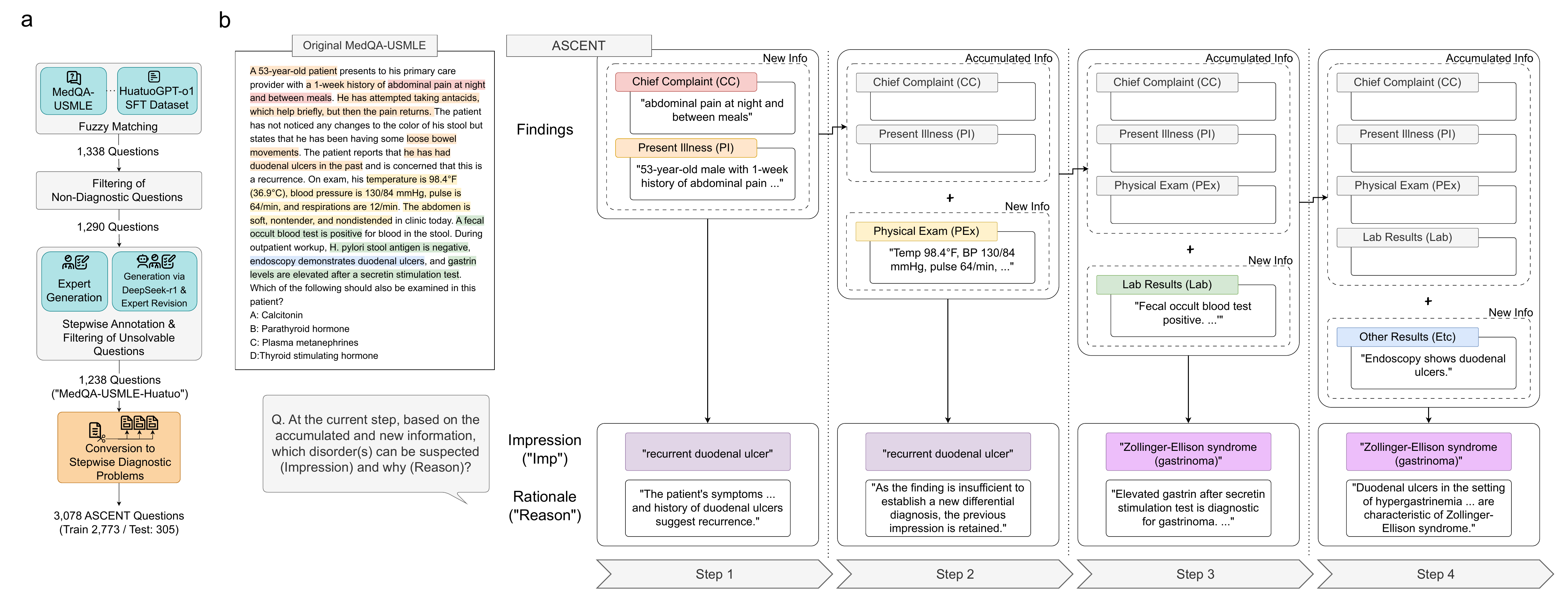}
\caption{\textbf{a}. Data construction workflow of ASCENT. Candidate questions from MedQA‑USMLE and the HuatuoGPT‑o1 SFT dataset were filtered and stepwise‑annotated by experts, either directly or via DeepSeek‑R1 with expert revision. The finalized questions were converted into stepwise diagnostic problems across clinical stages.
\textbf{b}. The structure of ASCENT data. At each step, clinical findings are paired with a list of suspected disorders (\textbf{Impression}) and the corresponding rationale (\textbf{Reason}). As steps progress and new clinical information accumulates, the Impression is refined or updated to reflect the evolving diagnostic reasoning. In evaluations with ASCENT, each step is treated as a separate question, and models are prompted to generate either only the Impression or both the Impression and the corresponding Reason.}
\label{fig1}
\end{figure*}

\section{ASCENT}
\subsection{Structured design for Stepwise Diagnosis}
\label{sec:dataset_structure}
Figure \ref{fig1}-b illustrates the stepwise structure of the ASCENT dataset, including the conversion of MedQA-USMLE questions into ASCENT data samples.
To emulate a more naturalistic flow of information in clinical settings, the content of each original MedQA-USMLE question was segmented and classified into one of the following EMR sections.

\begin{itemize}
  \item \textbf{Chief Complaint (CC)}
  briefly states the primary symptom that serves as the reason for seeking medical care, as reported by the patient.
  \item \textbf{Present Illness (PI)} provides a concise narrative of the patient's chief complaint, describing their onset, course, and associated features, while incorporating relevant contextual information such as age and gender.
  \item \textbf{Physical Examination (PEx)} consists of objective findings obtained by the clinician, typically through direct inspection, palpation, percussion, or auscultation.
  \item \textbf{Laboratory Results (Lab)} include outcomes from laboratory tests performed on biological specimens (e.g., blood, urine tests, or culture tests).
  \item \textbf{Imaging Results (Img)} present interpretative findings derived from radiological studies (e.g., X-ray, computed tomography (CT), or magnetic resonance imaging (MRI)).
  \item \textbf{Other Results (Etc)} encompass additional diagnostic and procedural findings not covered in the previously mentioned sections (e.g., electrocardiography (ECG), endoscopic evaluations, and biopsy results), as well as the post-treatment status and planned management.
\end{itemize}

Formally, each clinical case \( \mathcal{C} \) derived from a medical exam question is represented as an ordered sequence of steps:
\[
\mathcal{C} = [\mathbf{s}_0, \mathbf{s}_1, \dots, \mathbf{s}_n],
\]
where each step \( \mathbf{s}_t \) consists of:
\begin{itemize}
  \item \( f_t \): clinical findings newly acquired at step \( t \) corresponding to the information from one of the predefined EMR sections,
  \item \( i_t \): impression (suspected diagnosis or system) at step \( t \) based on the collected clinical information through previous and current steps \( f_0, \dots, f_t \),
  \item \( r_t \): rationale that supports the impression \( i_t \).
\end{itemize}

The first step \( \mathbf{s}_0 \), corresponding to the chief complaint (CC), includes only clinical findings \( f_0 \); both impression \( i_0 \) and rationale \( r_0 \) are defined as \texttt{None}, as this section is composed of briefly stated symptoms. As the CC step includes limited information, we construct segments only for steps \( t \geq 1 \). For each such step \( t \), a diagnostic problem is constructed as:

\begin{itemize}
  \item \textbf{Stepwise Diagnostic Context} \( X_t \): the accumulated findings, impressions, and rationales from steps \( 0 \) through \( t-1 \), along with the current step's findings:
  \[
  X_t = \left[ \bigcup_{i=0}^{t-1} \left( f_i,\ i_i,\ r_i,\ \right),\ f_t \right].
  \]
  \item \textbf{Target} \( Y_t \): depending on the task setup, the target can be one or more of the following: \( i_t \), \( r_t \).
\end{itemize}

This format ensures that the model predicts the impression (and/or the rationale) at each stage using only the information available up to that point, without access to the current step’s impression or rationale.

\subsection{Data Generation}
\label{sec:data_generation}
\paragraph{Problem Filtering.}
For direct comparison with HuatuoGPT-o1 and its Supervised Fine-Tuning (SFT) dataset, we retained only the MedQA-USMLE \citep{Jin2020MedQAUSMLE} problems also present in the HuatuoGPT-o1 SFT dataset, identified via fuzzy string matching with Python's FuzzyWuzzy library at a similarity threshold of 0.85.
To remove questions that do not require diagnostic reasoning, we filtered out those containing keywords related to research or experimental studies (e.g., ``experiment,'' ``epidemiologist,'' ``cohort'') and quantitative terms (e.g., ``probability''), followed by manual inspection. In total, 48 of the 1,338 questions were excluded by the filtering.
Among the filtered questions, simple knowledge-based questions that did not present clinical scenarios (e.g., items asking solely about pharmacology mechanisms or basic anatomy without a patient context) were excluded. In addition, questions requiring interpretation of the accompanying images were also omitted due to restricted accessibility to the images.
The resulting dataset containing 1,238 questions, referred to as ``MedQA‑USMLE‑Huatuo,'' was then subjected to stepwise annotation described in the following sections.

\paragraph{Stepwise Diagnostic Problems Generated by Experts.}
Four board‑certified physicians were recruited to annotate the MedQA‑USMLE‑Huatuo dataset into the ASCENT structure, which segments each case into sequential clinical steps and pairs the findings at each step with corresponding impressions and rationales.
Along with the dataset, the physicians were provided with the raw outputs from a Llama-3.1-8B-Instruct model fine-tuned on a separate in-house dataset in the same structure. They were guided to generate sequentially structured data in accordance with the dataset design.
After the initial annotation, two separate reviewers with medical expertise (a board‑certified physician, Y.H.K., and a medical student, T.K.), along with another reviewer (Y.C.) to address formatting and consistency, thoroughly reviewed and revised the outcome.
Overall, more than half of the dataset (680 questions) were annotated during this process; the corresponding edit distribution is reported in Appendix~\ref{apd:r1_provenance}.

\paragraph{DeepSeek-R1-Based Generation.}
The remaining part of the  MedQA-USMLE-Huatuo dataset was also used to generate the ASCENT dataset. Specifically, the DeepSeek-R1 model \citep{DeepSeekAI2025DeepSeekR1IR} was provided with a detailed instruction to classify the sections and generate the impression and rationales for 558 questions, based on the corresponding HuatuoGPT-o1 complex CoT.
The prompt template for DeepSeek-R1-based generation is shown in Appendix~\ref{apd:prompt}.

The generation outcomes were then meticulously reviewed and revised by two reviewers with medical expertise (Y.H.K., a board‑certified physician, and T.K., a medical student), along with another reviewer (Y.C.) to examine and correct formatting issues. All DeepSeek‑R1‑generated cases received independent review by Y.H.K. and T.K., with Y.H.K. thoroughly reviewing all of T.K.'s revisions; no case bypassed clinical review.

The DeepSeek-R1 drafts required substantially less revision than the Llama-3.1-8B drafts used for the expert-annotated subset: 57.1\% no edit and 17.5\% major edits (similarity $<$0.85) for DeepSeek-R1, versus 97.7\% major edits for the 680 Llama-3.1-8B drafts. Full provenance is reported in Appendix~\ref{apd:r1_provenance}.

\paragraph{Construction into Stepwise Diagnostic Problems.}
After annotation, each problem was segmented into multiple stepwise diagnostic problems, where each step consists of a Stepwise Diagnostic Context ($X_t$) and a corresponding Target ($Y_t$). If there existed a preceding Impression but the current Findings were not decisive enough to change it, the rationale was filled in with ``As the finding is insufficient to establish a new differential diagnosis, the previous impression is retained.'' The resulting ASCENT dataset comprises 3,078 questions. Both the expert‑generated and the DeepSeek‑R1‑generated, expert‑reviewed subsets were randomly divided at the level of the original MedQA-USMLE questions in a 9:1 training‑to‑test ratio (seed=3407), yielding 2,773 training and 305 test questions. Figure \ref{fig1}-a provides an overview of the data generation workflow.

\section{Experiment}
To evaluate the clinical reasoning capabilities of LLMs in stepwise diagnostic scenarios, we investigate how different training regimes affect performance on the ASCENT benchmark. We also examine the role of model scale, motivated by the observation that larger LLMs tend to exhibit stronger few-shot and CoT capabilities.

\subsection{Experimental Settings}
\paragraph{Models.} We employed a diverse set of LLMs spanning both open-source and proprietary models, as well as reasoning-augmented models for the comprehensive evaluation of stepwise reasoning abilities. Specifically, our experiments include Qwen2.5-7B-Instruct, Qwen2.5-32B-Instruct, Qwen3-8B, Qwen3-32B, HuatuoGPT-o1-7B, and DeepSeek-R1 (\textit{open-source models}), alongside OpenAI GPT-4o, o3-mini, and GPT-5.2 (\textit{closed-source models}). GPT-5.2 was included to assess potential saturation under a strong proprietary general-purpose model. Note that Qwen3-8B, Qwen3-32B, HuatuoGPT-o1-7B, DeepSeek-R1, OpenAI o3-mini, and GPT-5.2 were evaluated using their thinking modes, which are designed to enhance reasoning.

In addition, we fine‑tuned Qwen2.5‑7B‑Instruct and Qwen2.5‑32B‑Instruct on two counterpart datasets derived from the same MedQA‑USMLE questions: (1) the ASCENT training set and (2) the MedQA‑USMLE‑Huatuo dataset, which contains complex CoT data extracted from the HuatuoGPT‑o1 SFT dataset by matching with MedQA‑USMLE.
All models were fine-tuned for three epochs on eight NVIDIA H100 (80 GB) GPUs with a batch size of 16, using the AdamW optimizer and cosine learning rate scheduler with the learning rate of 1e-5, BF16 precision, and weight decay of 0.01.

\paragraph{Inference Settings.} The maximum generation length was set to 1,024 tokens for standard models and 4,096 tokens for reasoning models during the evaluation. This restriction of generated token length is based on the findings in \citet{Huang2025m1UT}, which report that excessively long generation lengths (i.e., over 4K tokens) can induce overthinking and degraded reasoning performance for problems in the medical domain. We further validate this choice via a token-length ablation in Appendix~\ref{apd:token_ablation}. Apart from the maximum generated token length, all experiments adopted the following inference settings for fair and reproducible comparison: temperature 0.6, top-P 0.95, top-K 20, presence penalty 1.0, frequency penalty 0.0, and minimum probability 0.0. All experiments were run on eight H100 GPUs, while proprietary models were evaluated via their commercial APIs.

\paragraph{Experimental Tasks for ASCENT.} As the ASCENT dataset provides two types of stepwise outputs—impressions and their supporting rationales—we conducted corresponding experiments on two tasks: (1) to generate impressions only (\textbf{\textit{Imp}}) and (2) to generate supporting rationales followed by impressions (\textbf{\textit{Imp + Reason}}). The prompt templates for \textit{Imp} and \textit{Imp + Reason} tasks can be found in Appendix~\ref{apd:prompt}.

\begin{table*}[p]
\centering
\setlength{\tabcolsep}{1.0mm}
\footnotesize
\begin{tabular}{clccccc}
\toprule
\textbf{Targets} & \textbf{Models}
& \textbf{Step 1}
& \textbf{Step 2}
& \textbf{Step 3+}
& \textbf{Overall}
& \textbf{95\% CI} \\
\midrule
\multirow{15}{*}{\textbf{Imp}} & \multicolumn{6}{l}{\textbf{Small Models ($<$10B)}} \\
\cmidrule(lr){2-7}
& Qwen2.5-7B-Instruct & 0.22 & 0.49 & 0.61 & 0.40 & [0.35,\,0.45] \\
& \quad + SFT w/ MedQA-USMLE-Huatuo & 0.23 & 0.53 & 0.65 & 0.43 & [0.38,\,0.48] \\
& \quad + SFT w/ ASCENT (ours) & \textbf{0.28} & \textbf{0.63} & \textbf{0.69} & \textbf{0.50} & [0.44,\,0.55] \\
\cmidrule(lr){2-7}
& HuatuoGPT-o1-7B & \textbf{0.33} & 0.59 & 0.69 & 0.50 & [0.45,\,0.55] \\
& Qwen3-8B & 0.29 & \textbf{0.62} & \textbf{0.73} & \textbf{0.51} & [0.45,\,0.56] \\
\cmidrule(r){2-7}

& \multicolumn{6}{l}{\textbf{Large Models ($\geq$10B)}} \\
\cmidrule(lr){2-7}
& OpenAI GPT-4o & 0.43 & 0.43 & 0.66 & 0.49 & [0.44,\,0.54] \\
& OpenAI GPT-5.2 & \textbf{0.47} & 0.56 & 0.57 & 0.53 & [0.48,\,0.57] \\
& Qwen2.5-32B-Instruct & 0.27 & 0.51 & 0.65 & 0.43 & [0.38,\,0.48] \\
& \quad + SFT w/ MedQA-USMLE-Huatuo & 0.27 & 0.48 & 0.68 & 0.43 & [0.38,\,0.49] \\
& \quad + SFT w/ ASCENT (ours) & 0.35 & \textbf{0.66} & \textbf{0.70} & \textbf{0.54} & [0.49,\,0.60] \\
\cmidrule(lr){2-7}
& Qwen3-32B & 0.35 & 0.59 & 0.62 & 0.49 & [0.44,\,0.54] \\
& DeepSeek-R1 & 0.56 & 0.70 & 0.75 & 0.65 & [0.60,\,0.70] \\
& OpenAI GPT-o3-mini & \textbf{0.66} & \textbf{0.74} & \textbf{0.76} & \textbf{0.72} & [0.67,\,0.76] \\

\midrule
\multirow{15}{*}{\textbf{Imp + Reason}} & \multicolumn{6}{l}{\textbf{Small Models ($<$10B)}} \\
\cmidrule(lr){2-7}
& Qwen2.5-7B-Instruct & 0.31 & 0.58 & \textbf{0.71} & 0.48 & [0.43,\,0.53] \\
& \quad + SFT w/ MedQA-USMLE-Huatuo & \textbf{0.34} & 0.56 & 0.70 & 0.50 & [0.44,\,0.55] \\
& \quad + SFT w/ ASCENT (ours) & 0.32 & \textbf{0.70} & 0.69 & \textbf{0.54} & [0.49,\,0.59] \\
\cmidrule(lr){2-7}
& HuatuoGPT-o1-7B & 0.37 & 0.63 & 0.71 & 0.54 & [0.48,\,0.59] \\
& Qwen3-8B & \textbf{0.41} & \textbf{0.68} & \textbf{0.78} & \textbf{0.59} & [0.54,\,0.65] \\
\cmidrule(r){2-7}

& \multicolumn{6}{l}{\textbf{Large Models ($\geq$10B)}} \\
\cmidrule(lr){2-7}
& OpenAI GPT-4o & 0.39 & 0.55 & 0.67 & 0.51 & [0.46,\,0.56] \\
& OpenAI GPT-5.2 & \textbf{0.53} & 0.62 & 0.58 & 0.57 & [0.52,\,0.62] \\
& Qwen2.5-32B-Instruct & 0.42 & 0.65 & 0.76 & 0.58 & [0.53,\,0.63] \\
& \quad + SFT w/ MedQA-USMLE-Huatuo & 0.39 & 0.63 & 0.76 & 0.56 & [0.51,\,0.61] \\
& \quad + SFT w/ ASCENT (ours) & 0.44 & \textbf{0.74} & \textbf{0.81} & \textbf{0.63} & [0.58,\,0.68] \\
\cmidrule(lr){2-7}
& Qwen3-32B & 0.44 & 0.66 & 0.76 & 0.58 & [0.53,\,0.63] \\
& DeepSeek-R1 & 0.51 & 0.70 & 0.74 & 0.63 & [0.58,\,0.68] \\
& OpenAI GPT-o3-mini & \textbf{0.58} & \textbf{0.75} & \textbf{0.77} & \textbf{0.69} & [0.64,\,0.74] \\

\bottomrule
\end{tabular}
\caption{Stepwise diagnosis performance of open‑source and proprietary models on the ASCENT test set ($n=305$), evaluated on two tasks: (1) generating impressions only (\textbf{\textit{Imp}}) and (2) generating supporting rationales followed by impressions (\textbf{\textit{Imp + Reason}}). Step columns report micro-F1 per stage (``Step3+'' aggregates Step~3 onward); the ``Overall'' column reports micro-F1 across all steps. The ``95\% CI'' column reports bootstrap 95\% confidence intervals on Overall micro-F1, computed via 10{,}000 test-record resamples (percentile method); under \textit{Imp}, the top reasoning cluster (o3-mini, DeepSeek-R1) is statistically separable from the lower cluster with non-overlapping CIs, whereas under \textit{Imp + Reason} the second-tier models cluster more closely and several CIs overlap (notably Qwen2.5-32B-Instruct + SFT w/ ASCENT and DeepSeek-R1). Models are grouped by size and whether they are specialized in reasoning, with bold numbers indicating the best result within each model category.}
\label{tab:stepwise_results}
\end{table*}

\subsection{Evaluation Metrics}
\label{sec:eval_metrics}

Model-generated impressions at each step are compared against the ground-truth impressions in ASCENT for evaluation. As both the predicted and ground-truth impressions can include multiple items, we use the micro-averaged F1 score as our evaluation metric.

For better understanding of model behavior under insufficient clinical context, we categorized impression outputs into four types: (1) \textbf{Correct Match} (e.g., ground truth: \textit{intrauterine growth restriction}; prediction: \textit{fetal growth restriction}), (2) \textbf{Overly Specific} (e.g., ground truth: \textit{respiratory disorder}; prediction: \textit{Chronic Obstructive Pulmonary Disease (COPD)}), (3) \textbf{Overly Broad} (e.g., ground truth: \textit{lupus nephritis}; prediction: \textit{glomerulonephritis}), and (4) \textbf{Others (Unrelated or Irrelevant)} (e.g., \textit{essential tremor}; prediction: \textit{Parkinson’s disease}), where the latter three indicate varying degrees of deviation from the ground‑truth impression. Model outputs were evaluated using OpenAI GPT‑4.1 (see Appendix~\ref{apd:prompt} for the full prompt; we additionally assess inter-judge agreement with GPT-5.2 in Appendix~\ref{apd:judge}), which classified each predicted impression into one of these categories. These labels were subsequently used to compute the F1 scores against the ground‑truth entities and to analyze failure cases.

\subsection{Results}
\paragraph{Main Results}

Table \ref{tab:stepwise_results} presents the micro-averaged F1 score for impressions.
Notably, the well-known LLMs, such as OpenAI's models and DeepSeek-R1, still underperform relative to their impressive results on the original MedQA-USMLE dataset. Models fine-tuned with conventional clinical CoTs, i.e., Qwen2.5 + SFT w/ MedQA-USMLE-Huatuo, also offer only marginal gains.

In contrast, models trained with ASCENT consistently outperformed their counterparts trained on the MedQA-USMLE-Huatuo dataset, highlighting the value of structured, stepwise supervision for diagnostic reasoning. For example, Qwen2.5-7B-Instruct fine-tuned on ASCENT achieved an F1 score of 0.50, surpassing the MedQA-USMLE-Huatuo counterpart (0.43). When prompted to generate rationales first (\textbf{\textit{Imp + Reason}}), the F1 score further improved to 0.54 versus 0.50 for the MedQA-USMLE-Huatuo version. These improvements reflect that the stepwise, incremental structure of ASCENT helps models draw candidate diagnoses from insufficient patient information, whereas models trained on the entire vignette show limited performance under stepwise evaluation.

Model size also contributed significantly to performance gains for Qwen2.5-Instruct models. Larger models consistently achieved higher F1 scores, especially with additional training on ASCENT. Qwen2.5-32B-Instruct fine-tuned on ASCENT reached an F1 score of 0.63 with rationale generation, clearly outperforming its 7B counterpart (0.54) with a large gap even compared to the other experiment cases. Across both 7B and 32B fine-tuning, F1 gains were driven primarily by precision: precision improved by 0.11--0.17 over pre-trained baselines while recall stayed essentially flat (Table~\ref{tab:diagnostic_results_full}, Appendix~\ref{apd:stepwise_perf}), indicating that ASCENT fine-tuning encourages more focused, clinically appropriate differential diagnoses rather than broader candidate enumeration. The Qwen3 series, however, violates this scaling pattern: Qwen3-32B underperforms Qwen3-8B due to consistently more \textit{Overly Broad} errors at every step (Appendix~\ref{apd:qwen3_scaling}). We hypothesize that the larger model's broader knowledge base produces more candidate disorders without a commensurate ability to narrow them under early-step uncertainty.

Reasoning-oriented “thinking models” outperformed their non-thinking counterparts even without task-specific fine-tuning, with the largest gains observed for OpenAI o3-mini and DeepSeek-R1; the Qwen3 series showed more modest improvements. OpenAI o3-mini achieved the highest overall F1 score (0.72). Interestingly, rationale generation (\textbf{\textit{Imp + Reason}}) did not consistently benefit all thinking models; while OpenAI o3-mini and DeepSeek-R1 exhibited slight drops (from 0.72 to 0.69 and from 0.65 to 0.63, respectively), Qwen3-8B and Qwen3-32B showed improvement with rationale generation (from 0.51 to 0.59 and from 0.49 to 0.58, respectively). This suggests that extended reasoning can sometimes induce overthinking or misalignment between reasoning and final answers.

Notably, even GPT-5.2—a strong proprietary general-purpose model—achieves only 0.53 / 0.57 (\textit{Imp} / \textit{Imp + Reason}), above GPT-4o (0.49 / 0.51) in point estimates but well below o3-mini in both modes (with non-overlapping 95\% CIs); the gap to DeepSeek-R1 is evident under \textit{Imp} (non-overlapping CIs), although their CIs overlap under \textit{Imp + Reason} (Table~\ref{tab:stepwise_results}). Reasoning‑specialized models consistently outperform stronger general‑purpose ones, suggesting that stepwise diagnostic reasoning rewards structured reasoning supervision more than raw capability, and that ASCENT remains far from saturation.

Overall, these findings demonstrate that ASCENT enhances diagnostic reasoning in both small and large models, scales effectively with model size, and narrows the gap with inherently reasoning‑oriented models, while also highlighting that the impact of rationale generation can vary across models.

\paragraph{Stepwise Analysis}
The diagnostic performance was further investigated in a stepwise manner, with the outcomes presented in the Step columns of Table \ref{tab:stepwise_results}.
We observe that, in contrast to human healthcare professionals, who can readily capture early clinical clues to infer candidate diagnoses, LLMs generally perform poorly in the early stages, where clinical evidence is sparse and uncertain, underscoring the need for sequential diagnostic training and intermediate-output evaluation. Since our setting is a mitigated scenario in which models receive ground-truth context, this gap is even more pronounced under realistic rollout conditions where errors propagate—an intuition confirmed by the rollout experiment in the Robustness Analysis section to follow.

Our results also indicate that training with structured, stepwise-annotated data can help address these shortcomings, particularly in the early stages. Models trained on ASCENT demonstrate improved accuracy at initial diagnostic steps, where less information is available, and \textit{Imp + Reason} training further improves later-stage performance (up to F1 0.81 at Step 3+ for ASCENT-trained Qwen2.5-32B-Instruct).

\begin{figure}[t]
\centering
\includegraphics[width=0.5\textwidth]{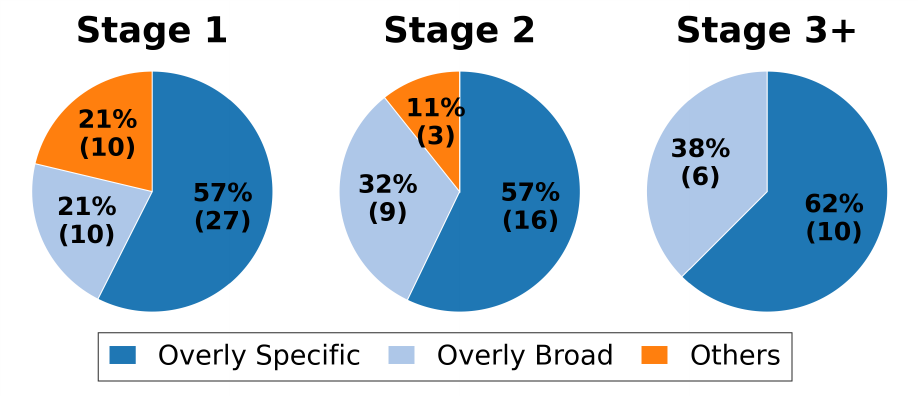}
\caption{Error type analysis for OpenAI o3‑mini on the \textit{Imp} task. Chart labels show the percentage of each error type, with corresponding case counts in parentheses.}
\label{fig_error_type}
\end{figure}

\begin{figure*}[t]
\centering
\includegraphics[width=1.0\textwidth]{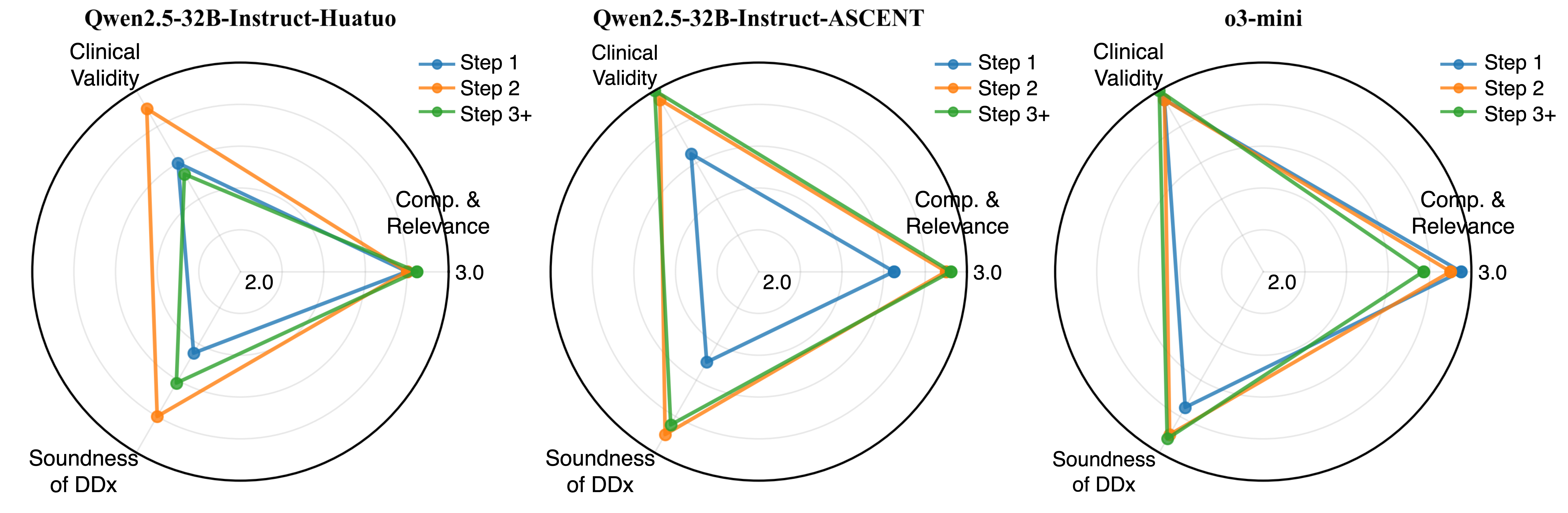}
\caption{Expert assessment of stepwise rationales. The quality of model-generated rationales was assessed on three clinician‑defined criteria: (1) \textbf{Comprehensiveness \& Relevance} (\textit{Comp. \& Relevance}), (2) \textbf{Clinical Validity}, and (3) \textbf{Soundness of Differential Diagnosis} (\textit{Soundness of DDx}). Solid lines indicate sequential diagnostic steps.}
\label{fig3}
\end{figure*}

\paragraph{Failure Case Analysis}
Figure \ref{fig_error_type} depicts the distribution of error types for the OpenAI o3-mini results on the \textbf{\textit{Imp}} task. Here, OpenAI o3‑mini was selected as the representative case because it achieved the best performance among the evaluated models, while other models exhibited similar error distribution trends.
Overall, the absolute number of errors in each category tends to decrease as more clinical context accumulates across steps.

This decline is most pronounced for the Others error type, where the sharp drop across diagnostic steps indicates that highly irrelevant predictions—those that fail to meaningfully leverage the given clinical context—become substantially less frequent as more information is obtained.
This trend further highlights the difficulty faced by LLMs in generating appropriate diagnoses when patient information is insufficient.
This finding may also be partly explained by the structure of the clinical scenarios in MedQA‑USMLE, where later‑stage information inherently reflects preceding outcomes because subsequent evaluations and tests are often selectively performed based on the diagnostic hypotheses formed in earlier stages.

\paragraph{Qualitative Evaluation on Model-Generated Rationales}

Beyond the model-judged error categories above, we additionally conducted a small-scale qualitative review of rationale quality with a clinician rater. The case study covers 20 problems, evaluating their generated rationales on a 3-point Likert scale based on three criteria: (1) \textbf{Comprehensiveness and Relevance}; (2) \textbf{Clinical Validity}; (3) \textbf{Soundness of Differential Diagnosis} (see Appendix~\ref{apd:rubric} for the detailed scoring rubric). A board-certified physician (Y.H.K.) developed the criteria and served as the rater. The rater, provided with the ground-truth rationales, remained blinded to the identities of the models. We treat this evaluation as indicative rather than conclusive given its limited scale.

Figure \ref{fig3} illustrates the stepwise assessment results for representative cases, while the full results are available in Appendix~\ref{apd:expert_eval_full}.
All models achieved average scores of 2.0 or higher across all reasoning steps and evaluation criteria. However, distinct patterns of stepwise performance were observed. Most models demonstrate performance gains as diagnostic steps progress, most notably when moving from Step 1 to Step 2. This jump likely stems from Step 1 including only patient‑provided information, whereas Step 2 typically incorporates more detailed observations from physical examinations.

Models trained on ASCENT follow a diagnosis‑centered approach, emphasizing targeted supporting evidence for suspected diagnoses. This style can result in lower early‑stage performance, particularly in Comprehensiveness and Relevance, as broad, nonspecific information in the early differential stages is often omitted. Similar trends were also observed in the model‑based assessment of impressions.

Intriguingly, models fine-tuned with MedQA‑USMLE‑Huatuo show generally lower and less stable performance, with a marked decline at Step 3+ in Clinical Validity and Soundness of Differential Diagnosis. This suggests that complex narrative CoTs from HuatuoGPT‑o1 alone may be insufficient to enhance stepwise diagnostic reasoning. A similar Step 3+ decline is also observed from other models in Appendix~\ref{apd:expert_eval_full}, although the contributing factors may differ between models.

\paragraph{Robustness Analysis.}
\label{sec:robustness}
The evaluation above isolates per-step reasoning by providing gold context up to each step (thus \textit{mitigated}). To characterize behaviour beyond this controlled setting—and to probe what ASCENT fine-tuning actually learns—we conducted four complementary analyses on o3-mini (strongest evaluated) and Qwen2.5-32B-Instruct fine-tuned on ASCENT (henceforth \textit{ours-32B}). We first present three analyses that support ASCENT-trained models' learned behaviour, then turn to a key caveat: error propagation under realistic rollout.

\textit{(1) Counterfactual perturbation.} On a stratified 75-case subset, we perturbed the most recent prior impression as \textit{overly broad}, \textit{overly specific}, \textit{irrelevant} (unrelated diagnosis), or \textit{order shift} (step swap). o3-mini was broadly resilient (largest drop: $-0.05$ F1 for \textit{overly specific}). ours-32B exhibited a striking asymmetry: collapse under \textit{overly broad} priors (F1\,=\,0.15) but robustness to \textit{irrelevant} (0.63), \textit{overly specific} (0.59), and \textit{order shift} (0.61) (full results in Appendix~\ref{apd:counterfactual}). We interpret this as the ASCENT-trained model treating prior impressions as an informative latent state summary: \textit{overly broad} priors carry too little signal, while \textit{irrelevant} ones are confidently incompatible and discarded.

\textit{(2) Format vs.\ content.} The MedQA-USMLE-Huatuo SFT baseline (Qwen2.5 fine-tuned on HuatuoGPT-o1 complex CoT data) serves as a \emph{format-matched control}: it produces step-level outputs in the same ASCENT-style format. Despite this identical output structure, it trails ASCENT fine-tuning by 0.04--0.11 F1 in Table~\ref{tab:stepwise_results}, indicating that matching the surface format alone cannot explain ASCENT's gains. Combined with the counterfactual asymmetry above, this rules out a pure style-transfer interpretation: a stylistic adapter would degrade comparably to general-purpose baselines under content perturbation, whereas ASCENT fine-tuning instead amplifies sensitivity to prior impression \emph{content}---consistent with the model learning to use accurate priors as diagnostic anchors.

\textit{(3) Judge-protocol robustness.} To rule out artifacts specific to a single LLM judge, we re-judged the full set of 313 impression-level prediction–reference pairs with the stronger GPT-5.2 model and compared against the original GPT-4.1 judgments. Inter-judge agreement was substantial under the 4-class scheme (Cohen's quadratic-weighted $\kappa = 0.78$) and remained substantial when \textit{Overly Specific} and \textit{Overly Broad} were collapsed into a single granularity-mismatch class ($\kappa = 0.76$); only 0.64\% of pairs reverse the dominant correct-vs-other direction, confirming that the F1 rankings reported above are stable across judge choice (full confusion matrices in Appendix~\ref{apd:judge}).

\textit{(4) Rollout evaluation.} Despite the positive evidence above, both evaluated models remain limited under realistic rollout, where each step receives the model's own prior prediction rather than gold context. For o3-mini, overall F1 collapses from 0.72 (mitigated) to 0.10 (rollout) for \textit{Imp} and from 0.69 to 0.09 for \textit{Imp+Reason}, with per-step degradation worsening monotonically from $-0.52$ at Step~1 to $-0.78$ at Step~3 (Appendix~\ref{apd:rollout}); ours-32B collapses even further, from 0.54 to 0.01 (\textit{Imp}) and 0.63 to 0.01 (\textit{Imp+Reason})---a larger \emph{relative} drop ($\sim$98\% vs $\sim$86\% for o3-mini) that reinforces the content-reliance interpretation above (a style-transfer adapter would be expected to degrade comparably to general-purpose baselines, not more). Strikingly, of 275 cases in which o3-mini produced a wrong impression, \textit{zero recovered} at any subsequent step (Table~\ref{tab:rollout_recovery})---indicating that error propagation is a fundamental obstacle for clinical deployment and motivating rollout-aware training as an important next step. While the rollout setting is closer to realistic deployment conditions, the mitigated setting remains informative on its own: it isolates per-step reasoning quality under controlled context and already exposes meaningful headroom on ASCENT, with rollout further compounding these gaps via error propagation.

\paragraph{Limitations and Future Work.} Building a dataset like ASCENT requires substantial human expert effort, underscoring the need for automated, reliable methods for generating stepwise diagnostic data. Moreover, as highlighted in prior work \citep{McDuff2025AMIE}, even with expert involvement, determining whether an impression is sufficiently specific can be subjective, which may lead to occasional disagreements over diagnostic granularity.

Due to limited resources, several complementary validations remain future work: (i)~a quantitative human baseline on the ASCENT test set; (ii)~a comprehensive inter-rater agreement analysis over the full dataset annotations, beyond the dual-reviewer revision protocol described in Section~\ref{sec:data_generation}; and (iii)~a comprehensive human review of LLM-as-a-Judge outputs to complement the model-vs-model judge agreement reported in Section~\ref{sec:robustness}.

Scaling ASCENT to larger sample sizes is also an important direction, both to support reliable per-chapter rate estimates and to broaden specialty coverage. Relatedly, although our automated ontology mapping showed promise (Appendix~\ref{apd:ontology}), finer-grained stratification of model errors along well-established medical ontologies is left for future work, and transfer to other sequential diagnostic datasets and to fully open-ended clinical settings is similarly not yet demonstrated. Finally, because ASCENT specifically targets stepwise diagnostic reasoning under EMR-aligned uncertainty, performance on ASCENT alone should not be interpreted as a comprehensive measure of clinical competence, which encompasses a variety of abilities such as patient communication, treatment planning, procedural skills, and longitudinal management.

\section{Conclusion}

We introduced ASCENT, a clinician-annotated benchmark that decomposes MedQA-USMLE vignettes into EMR-aligned diagnostic steps, making intermediate reasoning an explicit evaluation target—a regime that final-answer benchmarks cannot reach.

Our experiments yield the following salient findings. First, strong reasoning models substantially underperform their MedQA-USMLE accuracy on ASCENT, exposing a persistent gap between fully informed and stepwise diagnosis. Second, fine-tuning on ASCENT improves both small and large models over pre-trained and HuatuoGPT-o1 CoT-trained baselines, with gains driven primarily by precision—evidence that stepwise supervision teaches focused differential diagnosis rather than broader candidate enumeration. Third, robustness analyses reveal that ASCENT fine-tuning produces reliance on the content (not format) of prior impressions, while the strongest evaluated model (o3-mini) never recovers from a first wrong impression.

These findings call for rollout-aware training and for evaluation protocols that report both mitigated and rollout performance, since the two diverge sharply in our experiments. We release ASCENT as a public benchmark and substrate for stepwise clinical reasoning research.

\clearpage
\acks{We thank the physicians who annotated and the medical-expert reviewers who refined and qualitatively evaluated the ASCENT data. We are also grateful to the CHIL 2026 reviewers and area chair for feedback that shaped the robustness analyses.}

\bibliography{ascent}

\appendix

\setcounter{figure}{3}
\setcounter{table}{2}

\FloatBarrier
\clearpage

\section{Judge Reliability}
\label{apd:judge}
To assess the reliability of our LLM-as-a-Judge protocol, we re-ran the impression‑classification step using GPT‑5.2 as a stronger reference judge on the full test set, yielding 313 impression‑level pairs of judgments. Table~\ref{tab:judge_summary} summarizes agreement statistics; Tables~\ref{tab:judge_4class} and~\ref{tab:judge_3class} show the 4‑class and equiv‑23 (collapsed) confusion matrices.

\begin{table}[!h]
\centering
\small
\setlength{\tabcolsep}{1.0mm}
\begin{tabular}{lcccc}
\toprule
\textbf{Setting} & \textbf{Agree.} & \textbf{$\kappa$ unwt.} & \textbf{$\kappa$ lin} & \textbf{$\kappa$ quad} \\
\midrule
4‑class standard & 0.84 & 0.62 & 0.70 & 0.78 \\
3‑class equiv-23 & 0.85 & 0.63 & 0.69 & 0.76 \\
\bottomrule
\end{tabular}
\caption{Inter‑judge agreement between GPT‑4.1 (Judge A) and GPT‑5.2 (Judge B) on 313 impression‑level pairs. The equiv‑23 collapse merges \textit{Overly Specific} and \textit{Overly Broad} into a single granularity‑mismatch class. Both settings show substantial agreement under quadratic‑weighted Cohen's $\kappa$. The micro‑F1 of GPT‑5.2 was higher than GPT‑4.1 by $+0.08$, but only $0.64\%$ of pairs reverse direction (i.e., trend‑reversing disagreements), indicating that residual disagreements concentrate at class boundaries rather than systematic inversions.}
\label{tab:judge_summary}
\end{table}

\begin{table}[!h]
\centering
\small
\begin{tabular}{l|cccc}
\toprule
\textbf{A $\backslash$ B} & B=1 & B=2 & B=3 & B=4 \\
\midrule
\textbf{A=1} & 220 & 2 & 1 & 2 \\
\textbf{A=2} & 21 & 25 & 1 & 4 \\
\textbf{A=3} & 7 & 2 & 6 & 9 \\
\textbf{A=4} & 0 & 0 & 0 & 13 \\
\bottomrule
\end{tabular}
\caption{4‑class confusion matrix between GPT‑4.1 (rows, Judge A) and GPT‑5.2 (columns, Judge B). Class labels: 1 = Correct Match, 2 = Overly Specific, 3 = Overly Broad, 4 = Others. 95.9\% of disagreements fall within two ordinal classes ($|A-B|\le 2$); strictly adjacent disagreements ($|A-B|=1$, e.g., 1 vs.\ 2 or 2 vs.\ 3) account for 71.4\%.}
\label{tab:judge_4class}
\end{table}

\begin{table}[!h]
\centering
\small
\begin{tabular}{l|ccc}
\toprule
\textbf{A $\backslash$ B} & B=1 & B=2/3 & B=4 \\
\midrule
\textbf{A=1} & 220 & 3 & 2 \\
\textbf{A=2/3} & 28 & 34 & 13 \\
\textbf{A=4} & 0 & 0 & 13 \\
\bottomrule
\end{tabular}
\caption{Confusion matrix after collapsing \textit{Overly Specific} (2) and \textit{Overly Broad} (3) into a single granularity‑mismatch class.}
\label{tab:judge_3class}
\end{table}

\FloatBarrier

\section{Prompts for Data Generation, Inference, and Evaluation}\label{apd:prompt}
\subsection{Data Generation Prompt Template.}

\begin{figure*}[!htp]
\centering
\includegraphics[width=1.0\textwidth]{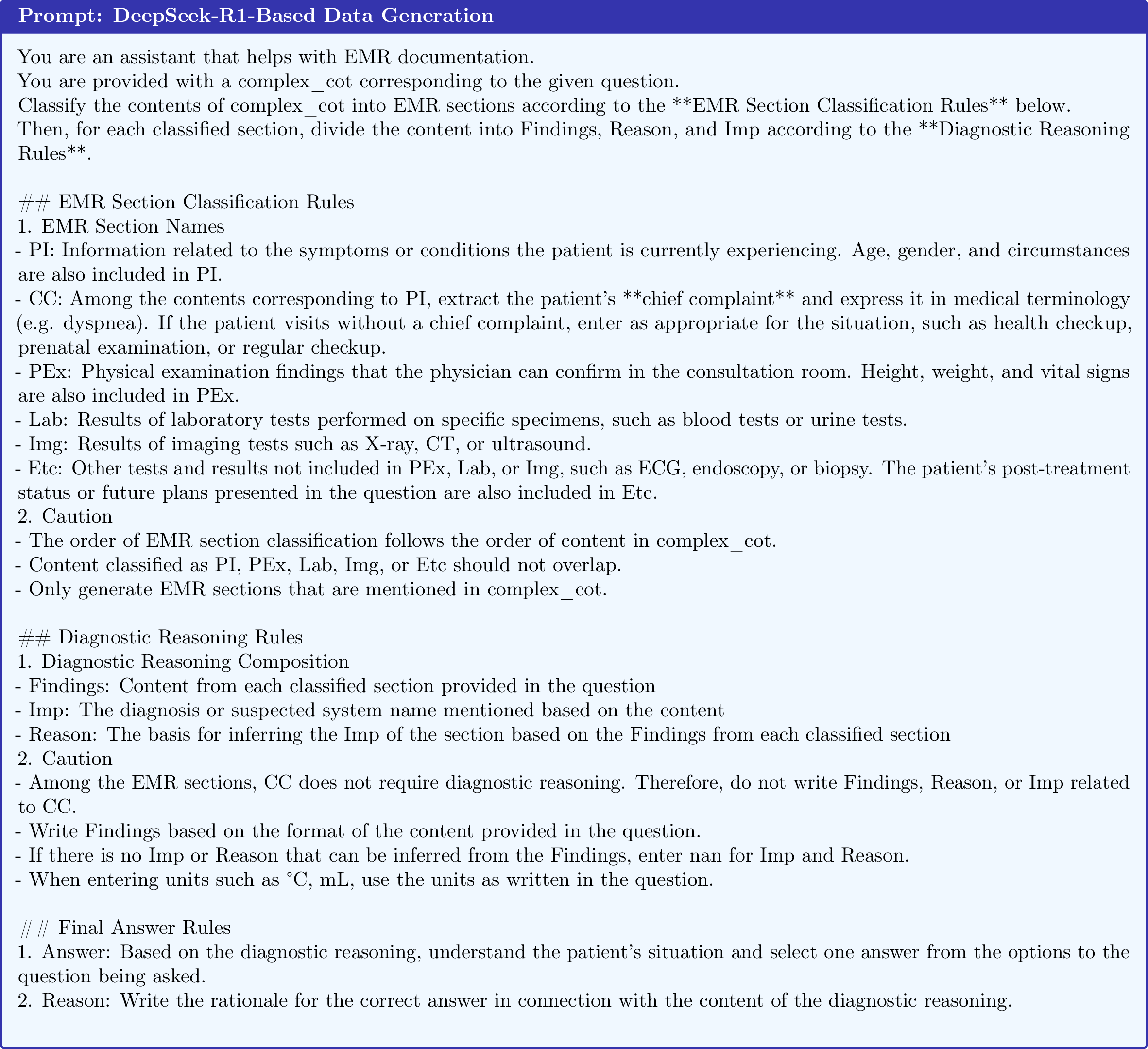}
\caption{The prompt template for DeepSeek-R1-based generation.}
\label{fig_prompt_r1gen_pt1}
\end{figure*}

\begin{figure*}[!htp]
\centering
\includegraphics[width=1.0\textwidth]{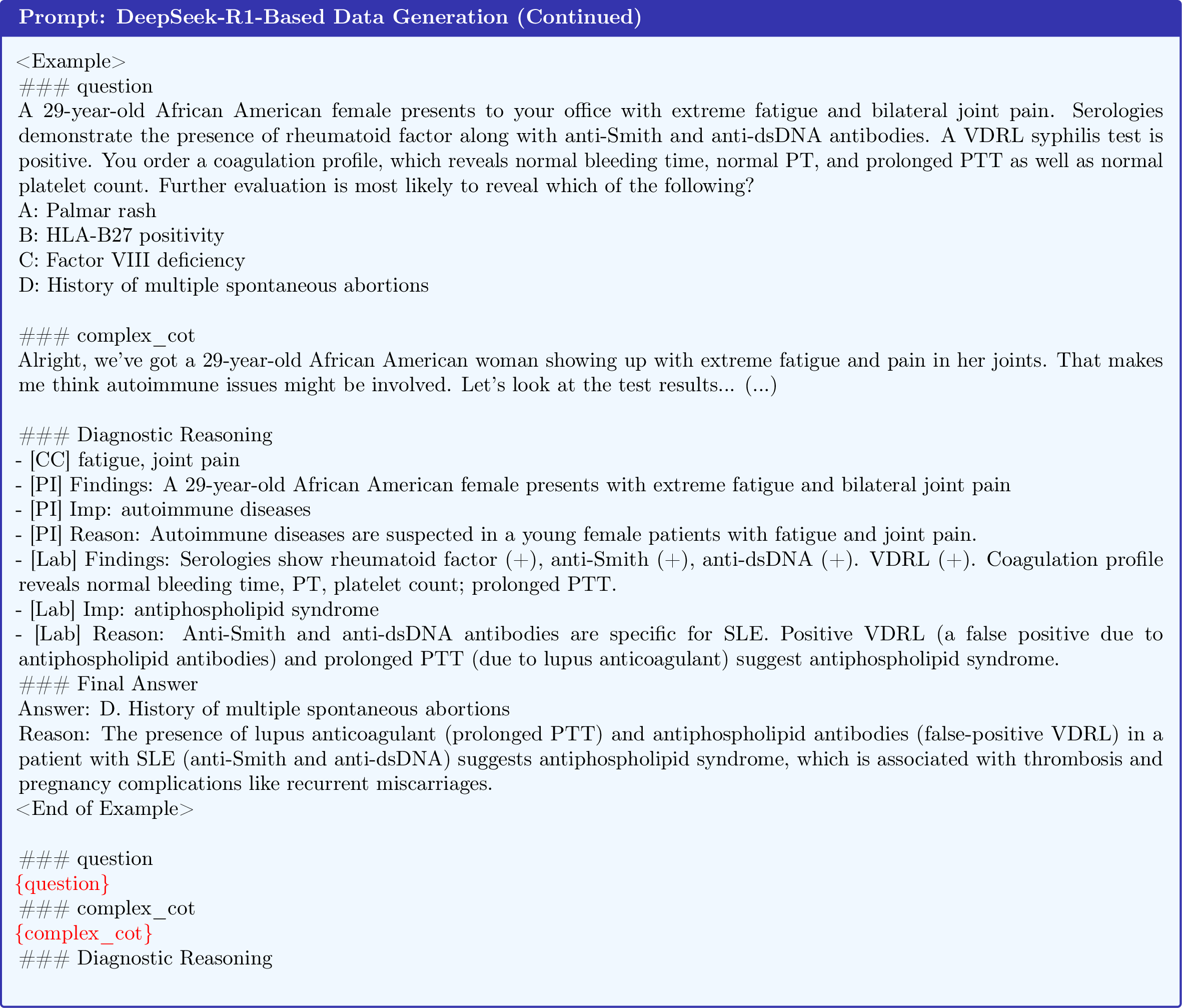}
\caption{The prompt template for DeepSeek-R1-based generation (continued).}
\label{fig_prompt_r1gen_pt2}
\end{figure*}

Figures~\ref{fig_prompt_r1gen_pt1} and ~\ref{fig_prompt_r1gen_pt2} show the prompt template used to generate ASCENT data with DeepSeek‑R1. The resulting outputs were meticulously reviewed and revised by reviewers prior to inclusion in ASCENT.

\subsection{Inference Prompt Template.}
\begin{figure*}[!htb]
\centering
\includegraphics[width=1.0\textwidth]{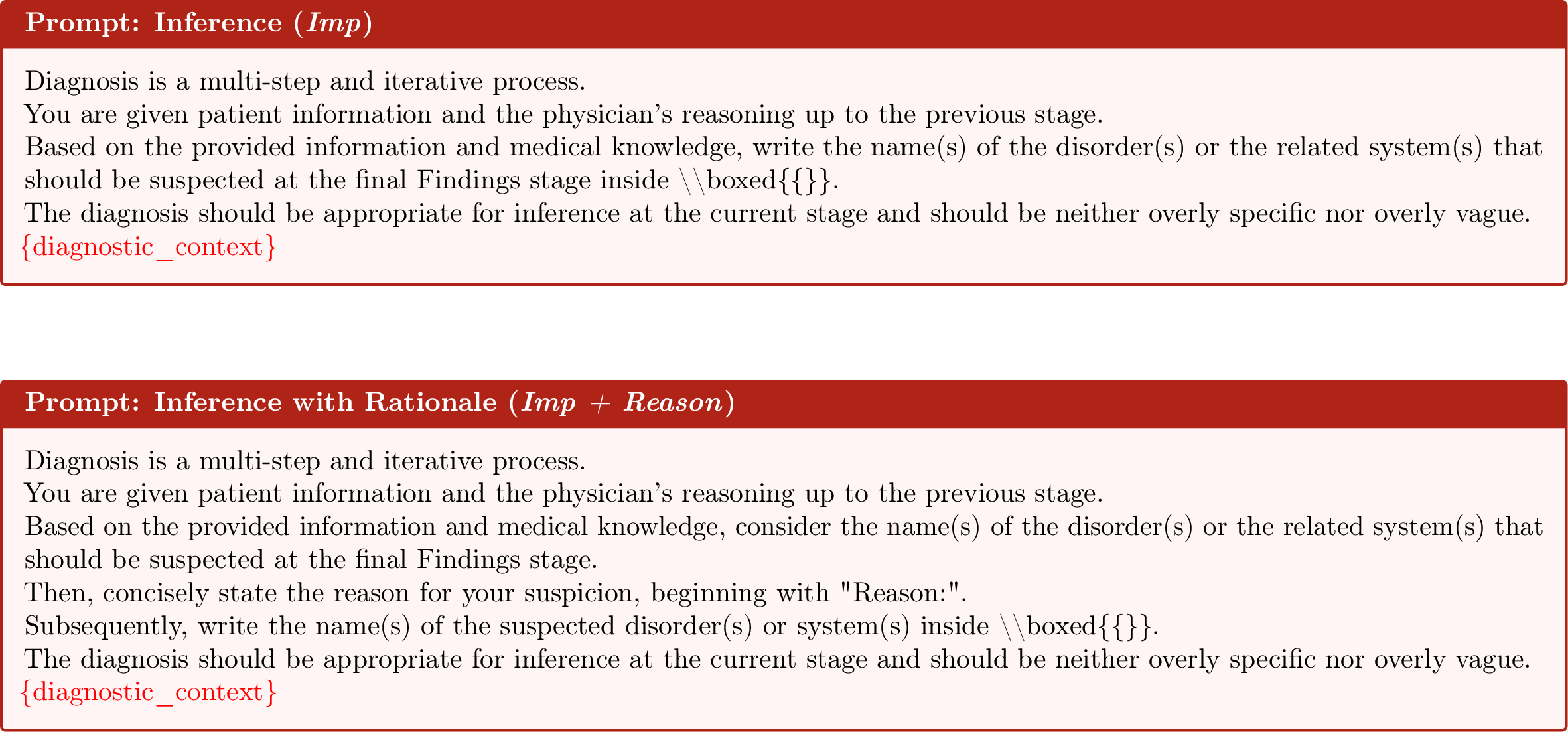}
\caption{The prompt template for inference for the impression (top: \textbf{\textit{Imp}}) and for the impression and rationale (bottom: \textbf{\textit{Imp + Reason}}).}
\label{fig_prompt_inference}
\end{figure*}

Figure~\ref{fig_prompt_inference} presents the prompt template for the impression (\textbf{\textit{Imp}}) and for the impression and rationale (\textbf{\textit{Imp + Reason}}) settings.

\subsection{Evaluation Prompt Template and Post‑Processing of Evaluation Results.}

\begin{figure*}[!htb]
\centering
\includegraphics[width=1.0\textwidth]{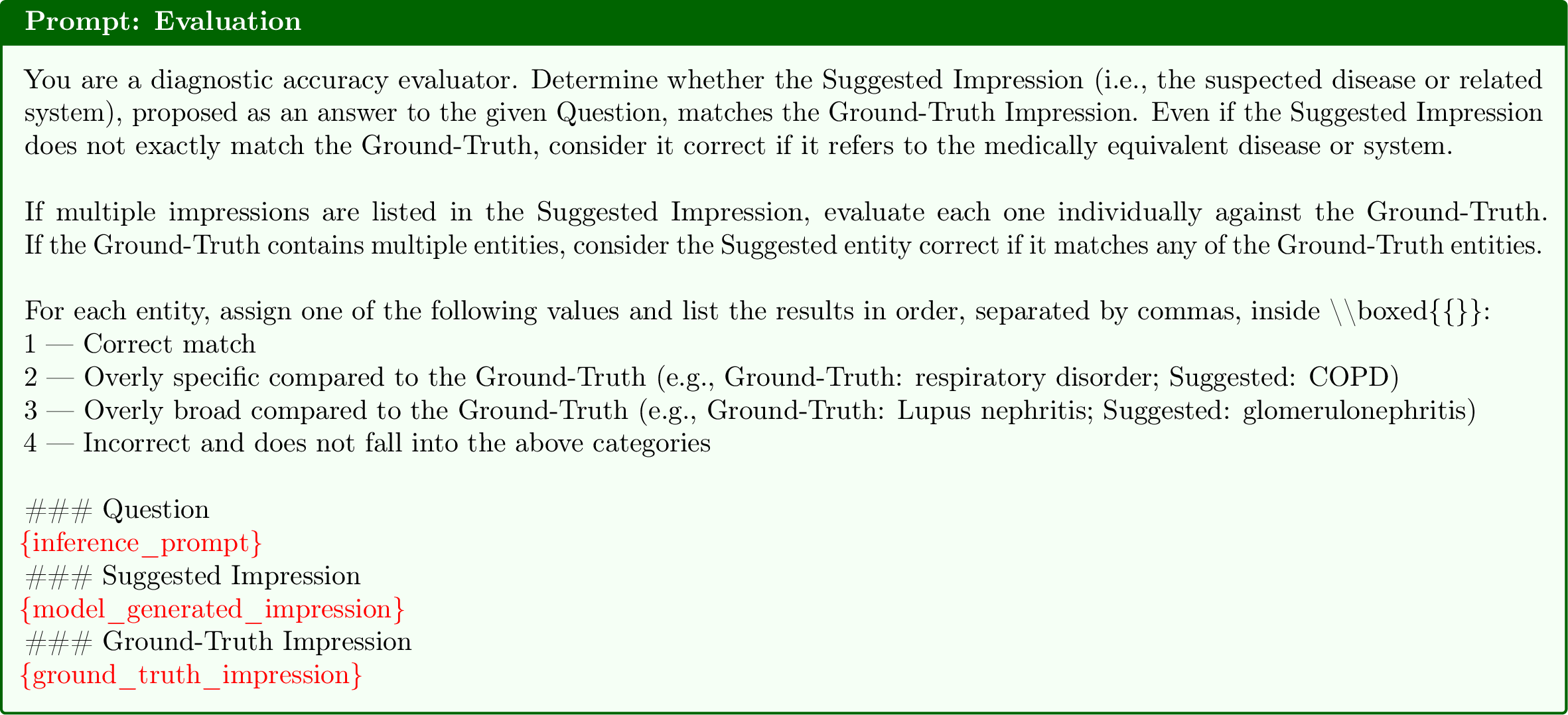}
\caption{The prompt template for model output evaluation.}
\label{fig_prompt_eval}
\end{figure*}

Figure~\ref{fig_prompt_eval} includes the prompt template used to evaluate and classify model-predicted impressions. The output was evaluated using OpenAI GPT-4.1 (temperature=0.0). To improve the consistency of the LLM-as-a-Judge approach, inference was repeated 5 times using the same prompt.

The resulting evaluation outputs were post‑processed to manually correct LLM‑induced formatting errors (e.g., missing \textbackslash\textbackslash boxed\{\{\}\}). We also handled cases where the number of evaluation outputs did not match the number of provided impressions by applying position‑wise majority voting to determine the final values. We then retained only the first k positions, where k equals the number of provided impressions, discarding any surplus values to maintain consistency of evaluation results.

\FloatBarrier
\clearpage

\begin{table*}[!htb]
\centering
\setlength{\tabcolsep}{1.0mm}
\footnotesize
\begin{tabular}{clccc}
\toprule
\textbf{Targets} & \textbf{Models}
& \textbf{Precision [95\% CI]}
& \textbf{Recall [95\% CI]}
& \textbf{F1 Score [95\% CI]} \\
\midrule
\multirow{15}{*}{\textbf{Imp}} & \multicolumn{4}{l}{\textbf{Small Models ($<$10B)}} \\
\cmidrule(lr){2-5}
& Qwen2.5-7B-Instruct & 0.33 [0.29,\,0.38] & 0.50 [0.44,\,0.55] & 0.40 [0.35,\,0.45] \\
& \quad + SFT w/ MedQA-USMLE-Huatuo & 0.38 [0.34,\,0.43] & 0.50 [0.44,\,0.55] & 0.43 [0.38,\,0.48] \\
& \quad + SFT w/ ASCENT (ours) & \textbf{0.50} [0.44,\,0.55] & \textbf{0.50} [0.45,\,0.55] & \textbf{0.50} [0.44,\,0.55] \\
\cmidrule(lr){2-5}
& HuatuoGPT-o1-7B & 0.46 [0.41,\,0.51] & \textbf{0.54} [0.49,\,0.60] & 0.50 [0.45,\,0.55] \\
& Qwen3-8B & \textbf{0.50} [0.45,\,0.56] & 0.51 [0.46,\,0.56] & \textbf{0.51} [0.45,\,0.56] \\
\cmidrule(r){2-5}

& \multicolumn{4}{l}{\textbf{Large Models ($\geq$10B)}} \\
\cmidrule(lr){2-5}
& OpenAI GPT-4o & 0.44 [0.40,\,0.49] & 0.54 [0.48,\,0.59] & 0.49 [0.44,\,0.54] \\
& OpenAI GPT-5.2 & 0.44 [0.39,\,0.49] & \textbf{0.65} [0.60,\,0.70] & 0.53 [0.48,\,0.57] \\
& Qwen2.5-32B-Instruct & 0.38 [0.33,\,0.43] & 0.51 [0.45,\,0.56] & 0.43 [0.38,\,0.48] \\
& \quad + SFT w/ MedQA-USMLE-Huatuo & 0.41 [0.35,\,0.46] & 0.47 [0.41,\,0.52] & 0.43 [0.38,\,0.49] \\
& \quad + SFT w/ ASCENT (ours) & \textbf{0.55} [0.49,\,0.60] & 0.54 [0.48,\,0.59] & \textbf{0.54} [0.49,\,0.60] \\
\cmidrule(lr){2-5}
& Qwen3-32B & 0.43 [0.38,\,0.48] & 0.57 [0.51,\,0.62] & 0.49 [0.44,\,0.54] \\
& DeepSeek-R1 & 0.59 [0.54,\,0.65] & \textbf{0.73} [0.68,\,0.78] & 0.65 [0.60,\,0.70] \\
& OpenAI GPT-o3-mini & \textbf{0.72} [0.67,\,0.77] & 0.71 [0.66,\,0.76] & \textbf{0.72} [0.67,\,0.76] \\

\midrule
\multirow{15}{*}{\textbf{Imp + Reason}} & \multicolumn{4}{l}{\textbf{Small Models ($<$10B)}} \\
\cmidrule(lr){2-5}
& Qwen2.5-7B-Instruct & 0.43 [0.38,\,0.48] & \textbf{0.55} [0.50,\,0.60] & 0.48 [0.43,\,0.53] \\
& \quad + SFT w/ MedQA-USMLE-Huatuo & 0.48 [0.43,\,0.54] & 0.52 [0.46,\,0.57] & 0.50 [0.44,\,0.55] \\
& \quad + SFT w/ ASCENT (ours) & \textbf{0.55} [0.50,\,0.61] & 0.53 [0.48,\,0.58] & \textbf{0.54} [0.49,\,0.59] \\
\cmidrule(lr){2-5}
& HuatuoGPT-o1-7B & 0.49 [0.43,\,0.54] & \textbf{0.59} [0.54,\,0.65] & 0.54 [0.48,\,0.59] \\
& Qwen3-8B & \textbf{0.61} [0.56,\,0.67] & 0.57 [0.52,\,0.62] & \textbf{0.59} [0.54,\,0.65] \\
\cmidrule(r){2-5}

& \multicolumn{4}{l}{\textbf{Large Models ($\geq$10B)}} \\
\cmidrule(lr){2-5}
& OpenAI GPT-4o & 0.48 [0.43,\,0.53] & 0.54 [0.49,\,0.60] & 0.51 [0.46,\,0.56] \\
& OpenAI GPT-5.2 & 0.50 [0.46,\,0.55] & \textbf{0.66} [0.61,\,0.72] & 0.57 [0.52,\,0.62] \\
& Qwen2.5-32B-Instruct & 0.54 [0.49,\,0.60] & 0.63 [0.57,\,0.68] & 0.58 [0.53,\,0.63] \\
& \quad + SFT w/ MedQA-USMLE-Huatuo & 0.54 [0.49,\,0.60] & 0.58 [0.52,\,0.63] & 0.56 [0.51,\,0.61] \\
& \quad + SFT w/ ASCENT (ours) & \textbf{0.65} [0.59,\,0.70] & 0.62 [0.57,\,0.67] & \textbf{0.63} [0.58,\,0.68] \\
\cmidrule(lr){2-5}
& Qwen3-32B & 0.55 [0.50,\,0.61] & 0.62 [0.57,\,0.67] & 0.58 [0.53,\,0.63] \\
& DeepSeek-R1 & 0.60 [0.55,\,0.66] & 0.67 [0.61,\,0.72] & 0.63 [0.58,\,0.68] \\
& OpenAI GPT-o3-mini & \textbf{0.69} [0.64,\,0.74] & \textbf{0.68} [0.63,\,0.73] & \textbf{0.69} [0.64,\,0.74] \\

\bottomrule
\end{tabular}
\caption{Overall diagnosis performance (micro precision, micro recall, and micro F1 score, with bootstrap 95\% CIs in brackets) of open‑source and proprietary models on the ASCENT test set ($n=305$), evaluated on two tasks: (1) generating impressions only (\textbf{\textit{Imp}}) and (2) generating supporting rationales followed by impressions (\textbf{\textit{Imp + Reason}}). CIs use 10{,}000 test-record bootstrap resamples (percentile method); within each resample, P, R, and F1 are computed from the same aggregated TP/FP/FN counts so the three values remain self-consistent. Bold marks the best value within each model category.}
\label{tab:diagnostic_results_full}
\end{table*}

\FloatBarrier
\section{Stepwise Diagnosis Performance}\label{apd:stepwise_perf}

\subsection{Overall diagnostic performance on the ASCENT test set.}
Table~\ref{tab:diagnostic_results_full} shows the overall diagnostic performance (micro precision, micro recall, and micro F1 score) of both open‑source and proprietary models on the ASCENT test set.

\subsection{A note on Qwen3 model scaling.}
\label{apd:qwen3_scaling}
As noted in the main manuscript, model size contributed to performance gains for Qwen2.5‑Instruct models. Larger models consistently achieved higher F1 scores, especially with additional training on ASCENT. However, Qwen3 models did not follow this trend, as the larger model did not outperform the smaller one. A closer inspection of stepwise errors, where a single question can produce multiple impressions and thus multiple errors, shows that Qwen3‑32B generated more overly broad errors across all steps (Step 1: 60 errors, Step 2: 32 errors, Step 3: 16 errors) than Qwen3‑8B (Step 1: 29 errors, Step 2: 25 errors, Step 3: 10 errors).

\subsection{Token‑length ablation.}
\label{apd:token_ablation}
We varied the maximum generation length for o3-mini from 512 to 4{,}096 tokens to assess sensitivity of stepwise diagnostic performance to the token budget. Table~\ref{tab:token_ablation} reports overall micro‑F1 across budgets for both \textit{Imp} and \textit{Imp+Reason}. Performance is essentially flat above 1{,}024 tokens: maximum variation is $0.03$ for \textit{Imp} and essentially flat for \textit{Imp+Reason} ($0.003$), well within one bootstrap standard error. Only the 512‑token setting showed non‑trivial degradation, and we observe no evidence of overthinking‑induced regression at higher budgets, supporting the choice of 4{,}096 tokens for reasoning models in our main experiments.

\begin{table}[!htb]
\centering
\small
\setlength{\tabcolsep}{2mm}
\begin{tabular}{lcc}
\toprule
\textbf{Token budget} & \textbf{Imp F1} & \textbf{Imp+Reason F1} \\
\midrule
512 & 0.69 & 0.65 \\
1{,}024 & 0.69 & 0.67 \\
2{,}048 & 0.69 & 0.67 \\
4{,}096 (default) & 0.72 & 0.67 \\
\bottomrule
\end{tabular}
\caption{Token‑length ablation for OpenAI o3-mini on the full ASCENT test set ($n=305$). Performance is approximately constant beyond 1{,}024 tokens.}
\label{tab:token_ablation}
\end{table}

\FloatBarrier

\subsection{Rollout (error propagation) evaluation.}
\label{apd:rollout}
We evaluated o3-mini under a rollout setting in which each step receives the model's own predicted impression from the prior step as context, rather than the gold impression used in the mitigated setting. Both task modes (\textit{Imp} and \textit{Imp+Reason}) were evaluated on the full ASCENT test set ($n=305$) using the standard GPT‑4.1 judge with five repeats and majority vote. Table~\ref{tab:rollout_step} reports per‑step F1 under both protocols, and Table~\ref{tab:rollout_recovery} reports the recovery rate after the first incorrect impression.

\begin{table}[!htb]
\centering
\small
\setlength{\tabcolsep}{1.0mm}
\begin{tabular}{lcccccc}
\toprule
& \multicolumn{3}{c}{\textbf{Imp}} & \multicolumn{3}{c}{\textbf{Imp+Reason}} \\
\cmidrule(lr){2-4}\cmidrule(lr){5-7}
 & Mit. & Roll. & $\Delta$ & Mit. & Roll. & $\Delta$ \\
\midrule
Step 1   & 0.66 & 0.14 & $-0.52$ & 0.58 & 0.12 & $-0.45$ \\
Step 2   & 0.74 & 0.10 & $-0.64$ & 0.75 & 0.08 & $-0.67$ \\
Step 3   & 0.82 & 0.03 & $-0.78$ & 0.83 & 0.06 & $-0.77$ \\
Step 4   & 0.62 & 0.00 & $-0.62$ & 0.62 & 0.07 & $-0.55$ \\
\midrule
Overall  & 0.72 & 0.10 & $-0.62$ & 0.69 & 0.09 & $-0.60$ \\
\bottomrule
\end{tabular}
\caption{Per-step and overall F1 for OpenAI o3-mini under \textbf{mitigated} (gold prior) and \textbf{rollout} (model's own prior) protocols (Step~5 omitted, $n=2$). Degradation worsens monotonically with step depth in both task modes. We focus the per-step analysis on o3-mini because it is the strongest model on ASCENT (Table~\ref{tab:stepwise_results}) and therefore offers the most informative test of error propagation: if even the best evaluated model cannot recover under rollout, the collapse is unlikely to be an artifact of weak baseline capability. The corresponding overall rollout collapse for ours-32B is reported in the main text.}
\label{tab:rollout_step}
\end{table}

\begin{table}[!htb]
\centering
\small
\setlength{\tabcolsep}{0.8mm}
\begin{tabular}{lccc}
\toprule
\textbf{Model} & \textbf{Total} & \textbf{First-wrong} & \textbf{Recovered} \\
\midrule
o3-mini (\textit{Imp}) & 305 & 275 & \textbf{0 (0.0\%)} \\
\bottomrule
\end{tabular}
\caption{Recovery rate after first incorrect impression under rollout. Of the 275 o3-mini cases (90.2\% of the test set) that produced a wrong impression at any step, none recovered at any subsequent step.}
\label{tab:rollout_recovery}
\end{table}

\subsection{Counterfactual perturbation.}
\label{apd:counterfactual}
On a stratified 75‑case subset, we perturbed the most recent prior impression with one of four variants: \textit{overly broad}, \textit{overly specific}, \textit{irrelevant} (an unrelated diagnosis), and \textit{order shift} (swapping two consecutive steps' prior impressions). Both o3-mini (the strongest evaluated model) and ours-32B (Qwen2.5-32B-Instruct fine‑tuned on ASCENT) were evaluated. Table~\ref{tab:counterfactual_full} reports the resulting F1 and resilience rate (fraction of cases producing at least one correct impression) for each variant.

\begin{table*}[t]
\centering
\small
\setlength{\tabcolsep}{1.5mm}
\begin{tabular}{llcc}
\toprule
\textbf{Model} & \textbf{Variant} & \textbf{F1} & \textbf{Resilience} \\
\midrule
\multirow{4}{*}{o3-mini}
 & overly broad     & 0.66 & 0.69 \\
 & overly specific  & 0.62 & 0.65 \\
 & irrelevant        & 0.65 & 0.68 \\
 & order shift      & 0.68 & 0.72 \\
\midrule
\multirow{4}{*}{ours-32B}
 & overly broad     & \textbf{0.15} & \textbf{0.16} \\
 & overly specific  & 0.59 & 0.63 \\
 & irrelevant        & 0.63 & 0.67 \\
 & order shift      & 0.61 & 0.64 \\
\bottomrule
\end{tabular}
\caption{Counterfactual perturbation of the most recent prior impression on a 75‑case subset. \textbf{Resilience} is an auxiliary metric we introduce for this analysis: the fraction of cases in which the model still produces at least one correct impression somewhere in the diagnostic trajectory under the perturbed prior, complementing the per-step F1 column. o3-mini (gold-prior baseline F1 on this subset = 0.67) is broadly resilient, with all four variants staying within roughly 5 F1 points of baseline. ours-32B exhibits a striking asymmetry: collapse under \textit{overly broad} priors (F1 = 0.15, far below its overall test-set F1 of 0.54 in Table~\ref{tab:diagnostic_results_full}) but robustness to \textit{irrelevant}, \textit{overly specific}, and \textit{order shift} variants, consistent with content‑based (rather than format‑based) reliance on the prior impression.}
\label{tab:counterfactual_full}
\end{table*}

\subsection{Draft revision analysis.}
\label{apd:r1_provenance}
To confirm that expert reviewers devoted comparable effort to revising and curating annotations from both draft sources—Llama-3.1-8B and DeepSeek-R1—we computed textual similarity between each draft and its final expert-revised version. A ``major edit'' corresponds to similarity $<0.85$. Table~\ref{tab:r1_provenance} reports the resulting edit distribution. The substantially lower edit rate for the DeepSeek-R1-drafted subset likely reflects DeepSeek-R1's stronger generation capacity relative to Llama-3.1-8B rather than reduced reviewer effort; all DeepSeek-R1 drafts received the same independent two-reviewer protocol described in Section~\ref{sec:dataset_structure}.

\begin{table*}[t]
\centering
\small
\begin{tabular}{lccc}
\toprule
\textbf{Draft source} & \textbf{No edit} & \textbf{Minor} & \textbf{Major} \\
\midrule
DeepSeek‑R1 ($n=558$) & 57.1\% & 25.5\% & 17.5\% \\
Llama‑3.1‑8B ($n=680$) & --- & 2.3\% & 97.7\% \\
\bottomrule
\end{tabular}
\caption{Edit distribution of physician revisions over the two draft sources used to construct ASCENT. ``Major'' is defined as similarity $<0.85$ between draft and final. The reviewer protocol is described in Section~\ref{sec:dataset_structure}.}
\label{tab:r1_provenance}
\end{table*}

\subsection{Ontology coverage.}
\label{apd:ontology}
As supporting evidence for the clinical groundedness of ASCENT's ground-truth impressions, we programmatically mapped all 1{,}152 entities to UMLS (2025AA), identifying 1{,}035 (89.8\%) matches. Of the matched entities, 836 (80.8\%) carry at least one SNOMED CT code, and 401 are linked through UMLS to ICD-10 codes spanning all 20 disease chapters (Table~\ref{tab:ontology_coverage}). This aggregate coverage suggests that ASCENT's impression vocabulary aligns well with standard medical ontologies and indicates potential utility for downstream tasks that benefit from standardized terminology; we leave full annotation of the entity-level mapping, which would warrant additional expert review, to future work.

\begin{table*}[t]
\centering
\small
\setlength{\tabcolsep}{1.5mm}
\begin{tabular}{lc}
\toprule
\textbf{Resource} & \textbf{Coverage} \\
\midrule
UMLS (2025AA)         & 1{,}035 / 1{,}152 (89.8\%) \\
SNOMED CT (via UMLS)  & 836 / 1{,}035 (80.8\%) \\
ICD‑10 (via UMLS)     & 401 entities, 20 chapters \\
\bottomrule
\end{tabular}
\caption{Ontology coverage of ASCENT ground‑truth impression entities. The 117 unmatched UMLS entities are free‑text terms. \textbf{``via UMLS''} indicates that mappings to SNOMED CT and ICD-10 were obtained indirectly, using each UMLS Concept Unique Identifier (CUI) as a bridge to the corresponding SNOMED CT and ICD-10 codes in the UMLS Metathesaurus (2025AA); we did not query SNOMED CT or ICD-10 directly. Consequently, SNOMED CT coverage is reported as a fraction of UMLS-matched entities, and the 401 ICD-10-linked entities span the full disease taxonomy (all 20 chapters).}
\label{tab:ontology_coverage}
\end{table*}

\section{Expert Evaluation Criteria and Scoring Rubric}\label{apd:rubric}

To further examine stepwise clinical reasoning, we conducted a qualitative case study on 20 problems. Model‑generated rationales were evaluated on a 3‑point Likert scale based on the following criteria: (1) \textbf{Comprehensiveness and Relevance}; (2) \textbf{Clinical Validity}; (3) \textbf{Soundness of Differential Diagnosis}. The detailed scoring rubric for the criteria is provided below.

\begin{itemize}
    \item \textbf{Comprehensiveness and Relevance} assesses whether all key relevant information is included and unnecessary details are excluded. \\
    3 — All key information is included and directly contributes to the differential diagnosis. \\
    2 — Some highly relevant information is missing, or some unnecessary details are present. \\
    1 — Important information is missing, or the majority of the content consists of irrelevant details.
    \item \textbf{Clinical Validity} evaluates whether the clinical content and medical knowledge are accurately and appropriately described. \\
    3 — All clinical information is accurate and aligns well with established medical knowledge. \\
    2 — Most information is accurate, but there are instances of incompleteness or ambiguity. \\
    1 — Contains inaccurate or misinterpreted clinical information.
    \item \textbf{Soundness of Differential Diagnosis} judges whether the resulting diagnosis is clinically appropriate and well-supported by the provided information. \\
    3 — The diagnosis is clear, well-supported, and clinically valid. \\
    2 — The diagnosis is somewhat plausible but either too broad or too specific. \\
    1 — The diagnosis is not clinically convincing or does not align with the information provided.
\end{itemize}

\section{Expert Evaluation on Model-Generated Rationales}\label{apd:expert_eval_full}
\begin{figure*}[!htb]
\centering
\includegraphics[width=1.0\textwidth]{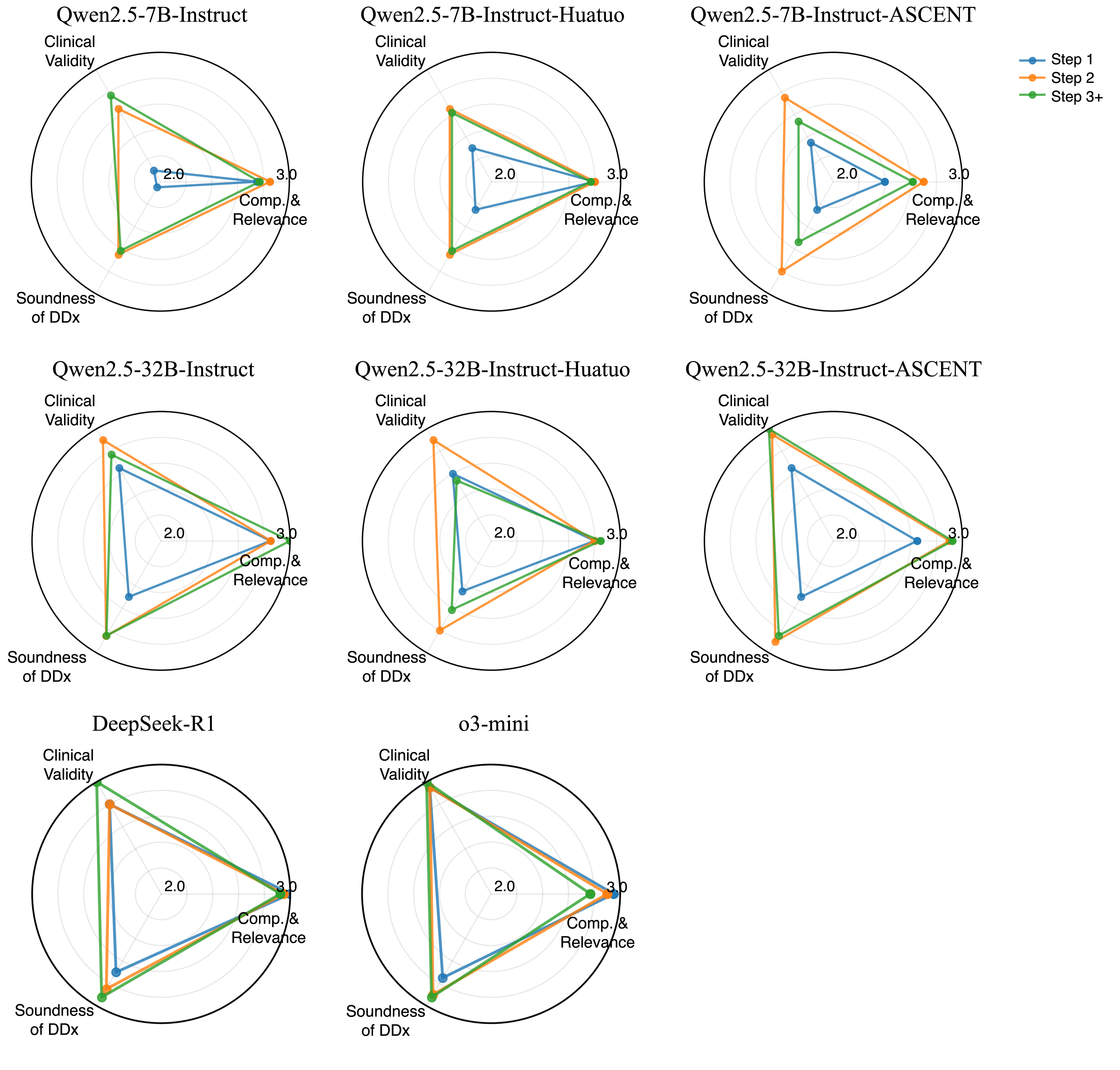}
\caption{Expert assessment of stepwise rationales (full results). The quality of model-generated rationales was assessed on three clinician‑defined criteria: (1) \textbf{Comprehensiveness \& Relevance} (\textit{Comp. \& Relevance}), (2) \textbf{Clinical Validity}, and (3) \textbf{Soundness of Differential Diagnosis} (\textit{Soundness of DDx}). Solid lines indicate sequential diagnostic steps.}
\label{fig_expert_eval_full}
\end{figure*}

Figure~\ref{fig_expert_eval_full} displays the full results of expert assessment on stepwise rationales.

\end{document}